# An Adam-enhanced Particle Swarm Optimizer for Latent Factor Analysis

Jia Chen
School of Cyber Science and Technology
Beihang University
Beijing, China
chenjia@buaa.edu.cn

Renyu Zhang
School of Cyber Science and Technology
Beihang University
Beijing, China
18373072@buaa.edu.cn

Yuanyi Liu*
School of Cyber Science and Technology
Beihang University
Beijing, China
YuanyiLiu@buaa.edu.cn

*Abstract*— Digging out the latent information from large-scale incomplete matrices is a key issue with challenges. The Latent Factor Analysis (LFA) model has been investigated in depth to analyze the latent information. Recently, Swarm Intelligence-related LFA models have been proposed and adopted widely to improve the optimization process of LFA with high efficiency, i.e., the Particle Swarm Optimization (PSO)-LFA model. However, the hyper-parameters of the PSO-LFA model have to tune manually, which is inconvenient for widely adoption and limits the learning rate as a fixed value. To address this issue, we propose an Adam-enhanced Hierarchical PSO-LFA model, which refines the latent factors with a sequential Adam-adjusting hyper-parameters PSO algorithm. First, we design the Adam incremental vector for a particle and construct the Adam-enhanced evolution process for particles. Second, we refine all the latent factors of the target matrix sequentially with our proposed Adam-enhanced PSO's process. The experimental results on four real datasets demonstrate that our proposed model achieves higher prediction accuracy with its peers.

## I. INTRODUCTION

With the smart applications and services becoming more complex and more widespread, there have generated and accumulated massive of High-Dimensional and Incomplete (HDI) data [1-7]. These data involve extremely useful latent information for analyzing the features of real entities and their relationships. Digging out the latent information from the HDI data is a challenging and interesting issue, which can help the smart applications and services provide better recommendation to the customers.

To analyze the HDI data effectively, the Latent Factor Analysis (LFA) model has been widely investigated [8-11]. The LFA model focuses on analyzing the latent factors of involved entities, which can represent the relationships between two entities and the latent characteristics of entities. The input data of an LFA model is generally an HDI matrix. The LFA model maps the target HDI matrix into a low-dimensional latent factor space, builds a low rank approximate matrix based on the already known HDI data, and resolves the unknown data in the approximate matrix via the iterative optimization process [12-17].

The iterative optimization process is quite important in LFA analysis, which is widely used together with optimization algorithms, such as the SGD, Adam, AdaGrad and AdaDelta etc. [18-21]. Among them, SGD is the most classical algorithm with high prediction accuracy. However, SGD algorithm needs to tune learning rate value for various datasets. To adjust the learning rate adaptively, the Adam, AdaGrad and AdaDelta algorithms are adopted recently. Although they obtain relatively high accuracy w/o tunning learning rate manually, much running time is needed to adjust the learning rate in each optimization iteration.

To further improve the optimization process, Luo et al.[22-25] introduce the Swarm Intelligence algorithms into the LFA model analysis. In this field, they have proposed a series of state-of-the-art models, i.e., the Particle Swarm Optimization (PSO)-LFA models, the Differential Evolution (DE)-LFA models, and the Beetle Antennae Search (BAS)-LFA models. Among them, the Hierarchical Position-transitional Particle Swarm Optimization-based LFA (HPL) model is an outstanding one, which optimizes the latent factors with two innovative PSO algorithms sequentially [24]. First, the Position-transitional Particle Swarm Optimization-based LFA (PLFA) model is adopted to optimize latent factors fast with adjusting the SGD's learning rate adaptively [22]. Second, the Mini-Batch PSO (MPSO) algorithm is adopted to refine the latent factors, which can achieve higher accuracy with acceptable additional time cost. However, the HPL model has to tune PSO's hyper-parameters for refining the latent factors of various datasets. That causes two problems. One is consuming lots of tunning time for various datasets, The other is the hyper-parameters can't be changed during the optimization process.

To resolve these two problems, we propose an Adam-enhanced HPL (ADHPL) model to refine the latent factors with adjusting the PSO's hyper-parameters adaptively. The proposed model first constructs a new gradient vector with the particle's velocity increment. Then, it adopts Adam to construct the first and second moment gradient vectors and update each particle with Adam algorithm. Last, the ADHPL model refines the latent factors with Adam-adjusting PSO algorithm. Our main contributions are listed as follows:

a) An ADHPL model. We incorporate the Adam-enhanced PSO evolution algorithm into the PSO's refinement process of HPL. Thereby, the proposed model refines all the latent factors w/o presetting the hyper-parameters. Compared with HPL, the proposed model obtains higher prediction accuracy with almost the same time cost.

b) An Adam-enhanced PSO's evolution algorithm. The classical PSO model updates each particle's velocity and position iteratively based on its velocity increment. We replace it with the designed Adam incremental vector and construct the Adam-enhanced evolution process for particles. The designed vector is based on the particle's velocity increment vector. Thereby, the proposed parametric adaptive evolution algorithm can simulate the evolution scheme of PSO effectively.

Section II describes the related definitions and algorithms. Section III proposes our innovative ADHPL model. Section IV describes experimental data and results. Finally, Section V concludes this paper and plans for the future work.

## II. Preliminaries

First, we summarize the mainly adopted symbols in Table I. Then we review the working mechanisms of an SGD-based LFA model, an ADAM algorithm and an HPL model.

### A. An LFA Model for HiDS Matrices

Let $Z^{|U|\times|I|}$ represents an HiDS matrix, which denotes the relationships between two large entity sets $U$ and $I$. Each element $z_{u,i} \in Z$ represents a specific relationship value between two specific entities $u \in U$ and $i \in I$. Besides, $Z_{un}$ denotes the unknown entity set of $Z$ while $Z_{kn}$ denotes the known entity set. For an HiDS matrix, the number of entities in $Z_{kn}$ is much smaller than in $Z_{un}$, $|Z_{kn}| \ll |Z_{un}|$.

To predict $Z_{un}$, an LFA model constructs a $F$-rank approximate matrix $\hat{Z}=PQ^T$, where $P^{|U|\times F}$ and $Q^{|I|\times F}$ denote the $F$-dimensional latent factor matrices for $U$ and $I$, respectively. Then, an objective function estimating the Euclidean distance between $Z$ and $\hat{Z}$ is given as [26-32], which is usually combined with linear biases for $P$ and $Q$, and an $L_2$-norm-based regularization item. The objective function is formulated as:

$$\varepsilon(P,Q,\mathbf{b},\mathbf{c}) = \frac{1}{2}\sum_{z_{u,i}\in Z_{kn}}\left(z_{u,i} - \sum_{f=1}^{F}p_{u,f}q_{f,i} - b_u - c_i\right)^2 \\ + \frac{\lambda}{2}\sum_{z_{u,i}\in Z_{kn}}\left(\sum_{f=1}^{F}p_{u,f}^2 + \sum_{f=1}^{F}q_{f,i}^2 + b_u^2 + c_i^2\right). \quad (1)$$

where $b_u$ and $c_i$ denote the specific bias vector for $U$ and $I$, $\lambda$ is regularization constant, respectively. Because (1) is non-convex and not analytically solvable, an LFA model commonly adopts SGD to optimize its latent factor set $\{P, Q, \mathbf{b}, \mathbf{c}\}$ with the preset gradient iteratively.

Recently, the PSO-related LFA models becomes popular, which incorporates the classical PSO algorithm into LFA's optimization process. Among them, the HPL model is an outstanding one. Thereby, we select it as the basic model.

### A. An HPL Model

An HPL model has a two-layer structure. The first layer adopts a PLFA model to pre-train latent factors, which adjusting SGD's learning rate by a P$^2$SO algorithm. Then the second layer refines latent factors with a Mini-batch PSO (MPSO) algorithm. It separates all the latent factors into $(|U|+|I|)$ mini-batch groups and optimize each group with PSO algorithm sequentially.

For each group $[\mathbf{p}_u,b_u]$, $\forall u \in |U|$, the MPSO algorithm constructs a swarm consisting of $S$ particles. Each particle's initial position vector $\mathbf{x}_s^u(0)$ consists of a vector nearby or equal to $[\mathbf{p}_u, b_u]$. Each particle's position $\mathbf{x}_s^u(\tau)$ and velocity vectors $\mathbf{v}_s^u(\tau)$ at the $\tau$-th iteration are represented as:

$$\mathbf{x}_s^u(\tau) = \left[\mathbf{p}_s^u(\tau), b_s^u(\tau)\right], \mathbf{v}_s^u(t) = \left[\dot{\mathbf{p}}_s^u(\tau), \dot{b}_s^u(\tau)\right], \tag{2}$$

where $\mathbf{p}_s^u(\tau)$, $b_s^u(\tau)$ denote the velocity of $\mathbf{p}_u$, $b_u$ at the $\tau$-th iteration for the $s$-th particle, $\dot{\mathbf{p}}_s^u(\tau)$ and $\dot{b}_s^u(\tau)$ denote their corresponding velocities. Then, the MPSO algorithm updates each $[\mathbf{p}_u, b_u]$ with traditional PSO and obtain the next iteration value. Velocity $\mathbf{v}_s^u(\tau+1)$ at the $(\tau+1)$-th iteration is updated as:

$$\begin{aligned}\mathbf{v}_s^u(\tau+1) &= \omega \cdot \mathbf{v}_s^u(\tau) + \gamma_1 r d_1 \left(\breve{\mathbf{h}}_s^u(\tau) - \mathbf{x}_s^u(\tau)\right) \\ &+ \gamma_2 r d_2 \left(\breve{\mathbf{g}}_u(\tau) - \mathbf{x}_s^u(\tau)\right) \\ &= \omega \left[\mathbf{p}_s^u(\tau), b_s^u(\tau)\right] + \gamma_1 r d_1 \left(\breve{\mathbf{h}}_s^u(\tau) - \left[\mathbf{p}_s^u(\tau), b_s^u(\tau)\right]\right) \\ &+ \gamma_2 r d_2 \left(\breve{\mathbf{g}}_u(\tau) - \left[\mathbf{p}_s^u(\tau), b_s^u(\tau)\right]\right),\end{aligned} \tag{3}$$

where $\gamma_1$ and $\gamma_2$ are the two acceleration coefficients. $rd_1$ and $rd_2$ are two random parameters between [0, 1], respectively. Then, the position $\mathbf{x}_s^u(\tau+1)$ of each $[\mathbf{p}_u, b_u]$ is updated as:

$$\mathbf{x}_s^u(\tau+1) = \mathbf{x}_s^u(\tau) + \mathbf{v}_s^u(\tau+1), \tag{4}$$

TABLE I. Mainly Adopted Symbols with Their Descriptions

| Symbol | Description |
|---|---|
| $\mathbf{b}, b_u$ | Bias vector of large entity set $U$, and its $u$-th element. |
| $b_s^u(\tau), c_s^u(\tau)$ | $s$-th particle's position with $\mathbf{b}$ and $\mathbf{c}$ at the $\tau$-th iteration. |
| $\dot{b}_s^u(\tau), \dot{c}_s^u(\tau)$ | $s$-th particle's velocity with $\mathbf{b}$ and $\mathbf{c}$ at the $\tau$-th iteration. |
| $\mathbf{c}, c_i$ | Bias vector of large entity set $I$, and its $i$-th element. |
| $F(\cdot)$ | Fitness function of the PSO algorithm. |
| $F, f$ | Max and active dimension number of the LF space. |
| $\breve{g}(\tau)$ | Best position vector at the $\tau$-th iteration in a particle swarm. |
| $\mathbf{h}_s$ | The $s$-th particle's historical best position vector. |
| $m(\tau)$ | The moving averages of gradient $\nabla\varepsilon(\tau)$. |
| $\hat{m}(\tau)$ | The corrected moving averages of gradient $\nabla\varepsilon(\tau)$. |
| $S, s$ | Max particle number in a particle swarm and the active one. |
| $P, p_{u,f}$ | An LF matrix for $U$, and its element in $u$-th row and $f$-column. |
| $\mathbf{p}_u$ | The $u$-th row vector of matrix $P$. |
| $\mathbf{p}_s^u(\tau), \mathbf{q}_s^u(\tau)$ | $s$-th particle's position with $\mathbf{p}_u$ and $\mathbf{q}_i$ at the $\tau$-th iteration. |
| $\dot{\mathbf{p}}_s^u(\tau), \dot{\mathbf{q}}_s^u(\tau)$ | $s$-th particle's velocity with $\mathbf{p}_u$ and $\mathbf{q}_i$ at the $\tau$-th iteration. |
| $Q, q_{f,i}$ | An LF matrix for $I$, and its element in $f$-th row and $i$-column. |
| $\mathbf{q}_i$ | The $i$-th row vector of matrix $Q$. |
| $T, \tau$ | Max iteration count of optimization algorithm, the current iteration. |
| $U, u$ | Large entity set one and its active element. |
| $I, i$ | Large entity set two and its active element. |
| $\mathbf{x}_s^u(\tau), \mathbf{v}_s^u(\tau)$ | $s$-th particle's position and velocity vector consisting of $[\mathbf{p}_u, b_u]$ at the $\tau$-th iteration. |
| $\mathbf{x}_s^i(\tau), \mathbf{v}_s^i(\tau)$ | $s$-th particle's position and velocity vector consisting of $[\mathbf{q}_i, c_i]$ at the $\tau$-th iteration. |
| $Z$, $z_{u,i}$ | An HiDS matrix consists of the relationship value for two entity sets $U$ and $I$, and each element in $Z$. |
| $Z_{kn}$, $Z_{un}$ | Known and unknown entry set in $Z$. |
| $\hat{Z}$, $\hat{z}_{u,i}$ | The approximation matrix of $Z$, and each element in $\hat{Z}$. |
| $\nabla \varepsilon_{u,i}(\cdot)$ | A partial derivative operator of the objective function to a specific relationship element between the $u$-th and the $i$-th entities. |
| $\Gamma$, $\Lambda$ | Training dataset and testing dataset of $Z$. |
| $\varepsilon()$ | Objective function estimating the distance between $Z$ and $\hat{Z}$. |
| $\eta$, $\lambda$ | Learning rate and regularization constant for a LFA model. |

To measure the accuracy of all the updated positions, the fitness function is constructed as:

$$F_1\left(x_s^u(\tau+1)\right) = \frac{1}{2} \sum_{z_{u,i} \in Z_{kn}} \left(z_{u,i} - \mathbf{p}_u q_i - b_u - c_i\right)^2 + \frac{\lambda}{2}\left(\|\mathbf{p}_u\|^2 + |b_u|^2\right). \tag{5}$$

With (5), each particle's updated fitness value is compared with its historical best one. The comparison formula at the $(\tau+1)$-th iteration is given as:

$$\breve{\mathbf{h}}_s^u(\tau+1) = \begin{cases} \mathbf{x}_s^u(\tau+1), & \text{if } F_1\left(\mathbf{x}_s^u(\tau+1)\right) < F_1\left(\breve{\mathbf{h}}_s^u(\tau)\right), \\ \breve{\mathbf{h}}_s^u(\tau), & \text{otherwise,} \end{cases} \tag{6}$$

Then, the global best position $\breve{g}^u(\tau+1)$ can be updated as:

$$\breve{\mathbf{g}}^u(\tau+1) = \underset{\breve{\mathbf{h}}_s^u(\tau+1)}{\arg\min}\left\{F_1\left(\breve{\mathbf{h}}_s^u(\tau+1)\right)\right\}. \tag{7}$$

With the above evolution process, the total $|U|$ $[\mathbf{p}_u, b_u]$ are refined sequentially to resolve the pre-mature problem in the PSO-LFA model. After all the $|U|$ mini-batch $[\mathbf{p}_u, b_u]$ are refined, the $|I|$ mini-batch $[\mathbf{q}_i, c_i]$ are refined sequentially. The HPL model achieves the balance of prediction accuracy and running efficiency.

*B. An Adam Algorithm*

The working mechanism of Adam is to adjust the instant learning rate at each iteration by calculating the ratio of moving averages of the gradient $\nabla \varepsilon(\tau)$ and the squared moving averages of the gradient $\nabla \varepsilon(\tau)^2$. Specifically, at the $(\tau+1)$-th iteration, the moving averages $m(\tau+1)$ and the squared moving averages $v(\tau+1)$ are calculated as:

$$\begin{cases} m(\tau+1) = \beta_1 m(\tau) + (1-\beta_1)\nabla \varepsilon(\tau), \\ v(\tau+1) = \beta_2 v(\tau) + (1-\beta_2)\nabla \varepsilon^2(\tau), \end{cases} \tag{8}$$

where $\beta_1$ and $\beta_2$ is the exponential decay rate parameter of $m$ and $v$ at the previous iteration, respectively. At the first iteration, $m(0)$ and $v(0)$ are initialized to 0 according to [18]. Besides, $\beta_1$ and $\beta_2$ are commonly set as 0.9 and 0.999 as in [18]. Then the corrected parameters $\hat{m}(\tau+1)$ and $\hat{v}(\tau+1)$ are formulated as:

$$\begin{cases} \hat{m}(\tau+1) = m(\tau+1)/(1-\beta_1), \\ \hat{v}(\tau+1) = v(\tau+1)/(1-\beta_2), \end{cases} \tag{9}$$

With the updated $\hat{m}(\tau+1)$ and $\hat{v}(\tau+1)$, the partial derivative operator $\nabla \varepsilon(\tau+1)$ can be updated as:

$$\nabla\varepsilon(\tau+1) = \alpha \cdot \hat{m}_a(\tau+1)/(\sqrt{\hat{v}_a(\tau+1)} + \psi), \qquad (10)$$

where $\alpha$ denotes the step-size setting, $\psi$ denotes a parameter avoiding denominator becoming zero. Then, each parameter in the objective function updates with $\nabla\varepsilon$ iteratively.

The above Adam algorithm has been widely adopted in deep learning, graph learning and HiDS matrices analysis. The Adam algorithm improves the accuracy w/o tunning the learning rate manually. Thereby, we incorporate Adam into an HPL model. The proposed ADHPL model does not only adjusts the hyper-parameters automatically, but also promotes the prediction accuracy.

### III. METHODOLOGY

An ADHPL model first reconstructs each particle's velocity updating formula with Adam, then apply the Adam-MPSO algorithm to refine the latent factors in the HiDS matrices. The specific working mechanism is introduced in detail as described the next two sub-sections.

*A. Construct the Adam incremental vector for a particle*

The first job is to construct a $(F+1)$-dimensional gradient vector $\nabla\varepsilon(\tau)$, which is the base incremental element for the Adam-adjusting velocity. To do so, we adopt the combination of the global best position $\breve{\mathbf{g}}^u(\tau)$ and each particle's best position $\breve{\mathbf{h}}_s^u(\tau)$ in (3) as the $(F+1)$-dimensional gradient vector $\nabla\varepsilon(\tau)$. The $\nabla\varepsilon(\tau)$ can be represented as:

$$\begin{aligned}
\nabla\varepsilon(\tau) &= rd_1(\breve{\mathbf{h}}^u(\tau) - \mathbf{x}_s^u(\tau)) + rd_2(\breve{\mathbf{g}}_s^u(\tau) - \mathbf{x}_s^u(\tau)) \\
&= rd_1\left(\breve{\mathbf{h}}^u(\tau) - \left[\mathbf{p}_s^u(\tau), b_s^u(\tau)\right]\right) \\
&\quad + rd_2\left(\breve{\mathbf{g}}^u(\tau) - \left[\mathbf{p}_s^u(\tau), b_s^u(\tau)\right]\right).
\end{aligned} \qquad (11)$$

Note that the constructed gradient vector $\nabla\varepsilon(\tau)$ equals to the incremental velocity vector between $\mathbf{v}_s^u(\tau+1)$ and $\mathbf{v}_s^u(\tau)$ in (3). Thereby, the $\nabla\varepsilon(\tau)$ can also represent the particle's velocity evolution during two adjacent iterations. Next, we build the first and second moment moving average parameters gradient vectors $\mathbf{m}(\tau+1)$ and $\mathbf{v}(\tau+1)$. The formulas can be reformulated from (8) as:

$$\begin{cases}
\mathbf{m}(\tau+1) = \beta_1 \mathbf{m}(\tau) + (1-\beta_1)\nabla\varepsilon(\tau) \\
\qquad = \beta_1 \mathbf{m}(\tau) + (1-\beta_1)\begin{bmatrix} rd_1\left(\breve{\mathbf{g}}^u(\tau) - \mathbf{x}_s^u(\tau)\right) \\ + rd_2\left(\breve{\mathbf{h}}_s^u(\tau) - \mathbf{x}_s^u(\tau)\right) \end{bmatrix}, \\
\mathbf{v}(\tau+1) = \beta_2 \mathbf{v}(\tau) + (1-\beta_2)\nabla\varepsilon_{u,i}^2(\tau) \\
\qquad = \beta_2 \mathbf{v}(\tau) + (1-\beta_2)\begin{bmatrix} rd_1\left(\breve{\mathbf{g}}^u(\tau) - \mathbf{x}_s^u(\tau)\right) \\ + rd_2\left(\breve{\mathbf{h}}_s^u(\tau) - \mathbf{x}_s^u(\tau)\right) \end{bmatrix}^2.
\end{cases} \qquad (12)$$

The corrected parameters $\hat{\mathbf{m}}(\tau+1)$ and $\hat{\mathbf{v}}(\tau+1)$ are given as:

$$\begin{cases}
\hat{\mathbf{m}}(\tau+1) = \mathbf{m}(\tau+1)/(1-\beta_1) = \beta_1 * \mathbf{m}(\tau)/(1-\beta_1) + \nabla\varepsilon(\tau) \\
\qquad = \beta_1 * \mathbf{m}(\tau)/(1-\beta_1) \\
\qquad\quad + \left(rd_1\left(\breve{\mathbf{g}}^u(\tau) - \mathbf{x}_s^u(\tau)\right) + rd_2\left(\breve{\mathbf{h}}^u(\tau) - \mathbf{x}_s^u(\tau)\right)\right), \\
\hat{\mathbf{v}}(\tau+1) = \mathbf{v}(\tau+1)/(1-\beta_2) = \beta_2 * \mathbf{v}(\tau)/(1-\beta_2) + \nabla\varepsilon^2(\tau) \\
\qquad = \beta_2 * \mathbf{v}(\tau)/(1-\beta_2) \\
\qquad\quad + \left(rd_1\left(\breve{\mathbf{g}}^u(\tau) - \mathbf{x}_s^u(\tau)\right) + rd_2\left(\breve{\mathbf{h}}^u(\tau) - \mathbf{x}_s^u(\tau)\right)\right)^2.
\end{cases} \qquad (13)$$

By adding (13) into (10), we reformulate the gradient $\nabla\varepsilon(\tau+1)$ in the next iteration with $\mathbf{m}(\tau)$, $\mathbf{v}(\tau)$ and $\nabla\varepsilon(\tau)$ as:

$$\begin{aligned}
\nabla\varepsilon(\tau+1) &= \alpha \cdot \frac{\beta_1 * \mathbf{m}(\tau)/(1-\beta_1) + \nabla\varepsilon(\tau)}{\sqrt{\beta_2 * \mathbf{v}(\tau)/(1-\beta_2) + \nabla\varepsilon^2(\tau)} + \psi} \\
&= \frac{\alpha \cdot \beta_1 \cdot \mathbf{m}(\tau)/(1-\beta_1) + \left(rd_1\left(\breve{\mathbf{g}}^u(\tau) - \mathbf{x}_s^u(\tau)\right) + rd_2\left(\breve{\mathbf{h}}^u(\tau) - \mathbf{x}_s^u(\tau)\right)\right)}{\sqrt{\beta_2 * \mathbf{v}(\tau)/(1-\beta_2) + \left(rd_1\left(\breve{\mathbf{g}}^u(\tau) - \mathbf{x}_s^u(\tau)\right) + rd_2\left(\breve{\mathbf{h}}^u(\tau) - \mathbf{x}_s^u(\tau)\right)\right)^2} + \psi}.
\end{aligned} \qquad (14)$$

With the above $(F+1)$-dimensional gradient vector $\nabla\varepsilon(\tau)$ and $\nabla\varepsilon(\tau+1)$ consists of particle swarm's global best position and each particle's best position, we can update the particles' velocity and positions with Adam. Besides, the constructed incremental vector uses the PSO's original incremental elements, which can remain the PSO's characteristics.

*B. Refine the latent factors in MPSO with Adam*

We construct ADHPL model by updating all the particles' positions with Adam in the MPSO algorithm. All the preset hyper-parameters are substituted with the Adam incremental vector. First, we construct and initialize the particle swarms for each mini-batch group $[\mathbf{p}_u, b_u]$, $\forall u \in |U|$ and $[\mathbf{q}_i, c_i]$, $\forall i \in |I|$ consisting of $S$ particles. For a specific $s$-th particle, the velocity $\mathbf{v}_s^u(0)$ and position $\mathbf{x}_s^u(0)$ are both initialized as a vector nearby the vector $[\mathbf{p}_u, b_u]$. Besides, the first and second moment moving average parameters $\mathbf{m}(0)$ and $\mathbf{v}(0)$ are preset. Second, we set velocity $\mathbf{v}_s^u(\tau+1)$ as $\nabla \varepsilon(\tau+1)$, which substitutes the formula in (3). We represent the new formula as:

$$\mathbf{v}_s^u(\tau+1) = \nabla \varepsilon(\tau+1) = \left[ \dot{\mathbf{p}}_s^u(\tau+1), \dot{b}_s^u(\tau+1) \right], \quad (15)$$

With the updated velocity $\mathbf{v}_s^u(\tau+1)$, the new position of the $s$-th particle $\mathbf{x}_s^u(\tau+1)$ is updated as:

$$\begin{aligned}\mathbf{x}_s^u(\tau+1) &= \mathbf{x}_s^u(\tau) + \mathbf{v}_s^u(\tau+1) = \mathbf{x}_s^u(\tau) + \nabla \varepsilon(\tau+1) \\ &= [p_s^u(\tau), b_s^u(\tau)] + \frac{\frac{\alpha \beta_1 \mathbf{m}(\tau)}{1-\beta_1} + \left( rd_1 \left( \breve{\mathbf{g}}^u(\tau) - \mathbf{x}_s^u(\tau) \right) + rd_2 \left( \breve{\mathbf{h}}^u(\tau) - \mathbf{x}_s^u(\tau) \right) \right)}{\sqrt{\frac{\beta_2 \mathbf{v}(\tau)}{1-\beta_2} + \left( rd_1 \left( \breve{\mathbf{g}}^u(\tau) - \mathbf{x}_s^u(\tau) \right) + rd_2 \left( \breve{\mathbf{h}}^u(\tau) - \mathbf{x}_s^u(\tau) \right) \right)^2} + \psi}.\end{aligned} \quad (16)$$

Based on the above update scheme, each $[\mathbf{p}_u, b_u]$ is updated with Adam until reaches the convergence conditions. After refining all the $|U|$ $[\mathbf{p}_u, b_u]$, the ADHPL model continues refining all the $|I|$ $[\mathbf{q}_i, c_i]$ until reaches the convergence conditions.

*C. Algorithm Design and Analysis*

With the above latent factor refining process, we can construct an ADHPL model. It can be seen as a sequential Adam-PSO refining process for all the latent factors in the target HDI matrix. The workflow of the ADHPL model is depicted in Fig. 1.

First, all the pre-trained latent factors are grouped by rows and columns. Thereby, all the latent factors are separated as the $|U|$ row groups and $|I|$ column groups. Each row group contains a latent factor pair $[\mathbf{p}_u, b_u]$, and each column group contains a latent factor pair $[\mathbf{q}_i, c_i]$.

Second, for each $[\mathbf{p}_u, b_u]$, $\forall u \leq |U|$, we perform the Adam-PSO to refine the latent factors while fixing the others. Each refined latent factor is transferred back to $P$, $\mathbf{b}$.

Third, for each $[\mathbf{q}_i, c_i]$, $\forall i \leq |I|$, we perform the Adam-PSO to refine the latent factors while fixing the others. Each refined latent factor is transferred back to $Q$, $\mathbf{c}$.

Finally, after all the groups in $\{P, Q, \mathbf{b}, \mathbf{c}\}$ have been refined on training dataset once, calculate RMSE value on validation dataset. If the model converges on validation dataset, stop the refinement process and evaluate its RMSE value on testing dataset, otherwise continue refinement.

Based on the above steps, the proposed ADHPL model can refine all the latent factors in a target HiDS matrix sequentially with Adam-adjusting PSO algorithm.

IV. EXPERIMENTAL RESULTS AND ANALYSIS

*A. General Settings*

**Evaluation Protocol.** To compare our proposed model's prediction accuracy and efficiency with some state-of-the-art models, we choose RMSE to measure the prediction accuracy. We choose the CPU running time cost to measure the models' efficiency. We perform all the experiments on a workstation that has a 32GB RAM and a 3.91 GHz Xeon CPU. The formulas of RMSE are given as:

$$RMSE = \sqrt{\frac{1}{|\Lambda|} \left( \sum_{r_{u,i} \in \Lambda} (r_{u,i} - \hat{r}_{u,i})^2 \right)}.$$

where $\Lambda$ represents the testing set disjoint with the training and validation sets.

**Dataset.** We adopt four datasets ML10M[33], Flixster[34],

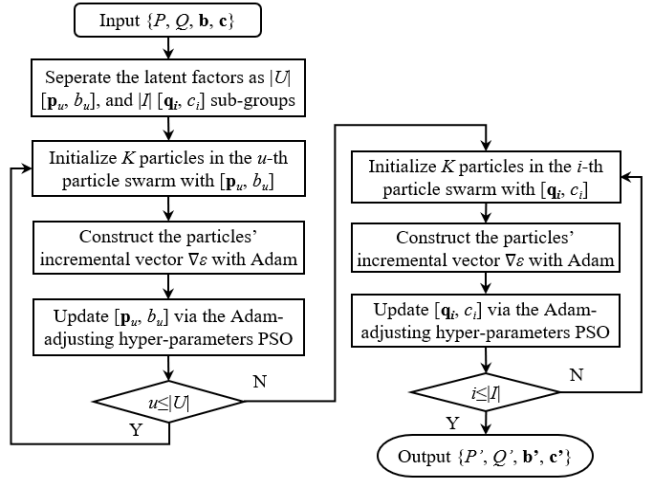

Fig.1 Flow Chart of the ADHPL Model

Douban[35] and ExtEpinion[36] to perform our models. We divide each dataset as training, validation and testing sub-datasets, which are 70%, 10% and 20%, respectively.

**Model Settings.** We adopt the SGD-based LFA, Adam-based LFA, PLFA with biases, $A^2$BAS-PLFA[37], HPL and SGDE-PLFA models as the comparison models. The dimension of LF space $f$ is set as 20. The regularization coefficient is preset the same value for all the involved models. All the other hyper-parameters are set as the original settings.

*B. Performance comparison*

The lowest RMSE value and Friedman rank of all the involved models on four datasets are given in Table II. The time cost is given in Table III.

*1) Comparison of prediction accuracy*

From Table II, we summarize the ADHPL model achieves relatively higher prediction accuracy compared with its peers. Two specific conclusions can be summarized as follows:

a) The Adam-HPL's prediction accuracy evaluated by RMSE is the highest of all the involved models, except for the SGDE-PLFA. For instance, the ADHPL's RMSE on ExtEpinion equals to 0.5279, which is 0.25% lower than SGDE-PLFA's 0.5292, 1.16% lower than HPL's 0.5341, and even much lower than other models. The ADHPL's RMSE on Flixster equals to 0.8599, which is 0.13% lower than SGDE-PLFA's 0.8610, 0.19% lower than $A^2$BAS-PLFA's 0.8615, and much lower than other models.

One exception is the SGDE-PLFA model. Our proposed ADHPL outperforms two datasets compared with SGDE-PLFA, performs equally well on one dataset, and performs worse on one dataset. For instance, on ML10M, the ADHPL's RMSE value equals to 0.7840, which is 0.04% higher than the SGDE-PLFA's 0.7837, while this value is lower than all the others. Besides, on Douban, the ADHPL performs as well as SGDE-PLFA, while they all outperform the others.

b) The ADHPL's prediction accuracy gain is statistically significant compared with its peers. The Win/Loss results are recorded in the next-to-last row of Table II, from which we can observe that the ADHPL and SGDE-PLFA both

TABLE II. COMPARISON RESULTS IN RMSE, INCLUDING WIN/LOSS COUNTS STATISTIC AND FRIEDMAN TEST, WHERE * INDICATES THE MODEL HAS A LOWERE RMSE THAN THE COMPARISON MODELS.

| Test Cases | SGD-based LFA | Adam-based LFA | PLFA with Biases | $A^2$BAS-PLFA | SGDE-PLFA | HPL | ADHPL |
|---|---|---|---|---|---|---|---|
| ExtEpinion | 0.7358 | 0.7342 | 0.5361 | 0.5327 | 0.5292 | 0.5341 | **0.5279*** |
| Flixster | 0.9433 | 0.9464 | 0.8720 | 0.8615 | 0.8610 | 0.8656 | **0.8599*** |
| Douban | 0.7176 | 0.7213 | 0.7076 | 0.7063 | **0.7027*** | 0.7028 | **0.7027*** |
| ML10M | 0.7872 | 0.7901 | 0.7854 | 0.7843 | **0.7837*** | 0.7850 | 0.7840 |
| Win/Loss | 4/0 | 4/0 | 4/0 | 4/0 | 3/1 | 4/0 | -- |
| F-rank | 6.25 | 6.75 | 5.0 | 3.25 | **1.5** | 3.75 | **1.5** |

* A lower Friedman rank value indicates higher rating prediction accuracy.

TABLE III. COMPARISON RESULTS IN TIME COST

| Dataset | SGD-based LFA | Adam-based LFA | PLFA with Biases | $A^2$BAS PLFA | SGDE-PLFA | HPL | Adam HPL |
|---|---|---|---|---|---|---|---|
| ML10M | 874 | 891 | 157 | **115** | 280 | 249 | 235 |
| ExtEpinion | 1152 | 4331 | **369** | 1462 | 700 | 438 | 426 |
| Flixster | 451 | 3393 | **91** | 331 | 212 | 222 | 252 |
| Douban | 1193 | 2064 | 340 | **298** | 541 | 476 | 482 |

win the others on all the datasets. The Friedman test results are recorded in the last row of Table II. Both ADHPL and SGDE-PLFA's F-rank value equal to 1.5, which is much less than other models.

*2) Comparison of computational efficiency*

From Table III, we summarize three conclusions of the ADHPL model's time cost:

a) It costs much less than the SGD-based LFA and Adam-based LFA. For instance, the ADHPL costs only 482s on Douban, which is 59.6% lower than SGD-based LFA's 1193s, 76.65% lower than Adam-based LFA's 2064s.

b) It costs about the same amount of time with other Swarm Intelligence-related LFA models, i.e., $A^2$BAS-PLFA, HPL and SGDE-PLFA. Furthermore, they all cost considerably more time than PLFA model. The reason is that they all refine the latent factors after the PLFA model convergences. However, the extra time of about a hundred seconds is worthy for promoting the prediction accuracy.

According to the above analysis, we conclude that the ADHPL model obtains superiority in terms of prediction accuracy, while convergences relatively fast.

V. CONCLUSIONS

This paper focuses on refining the latent factors with an Adam-enhanced HPL model. In this model, we design a base gradient with particle's velocity increment. With the proposed gradient, we propose an Adam-PSO algorithm, which can remain the characteristics of PSO w/o tunning PSO's hyper-parameters. Using this algorithm, we construct the ADHPL model to refine the latent factors sequentially with row and column sub-groups. The experimental results from four industrial datasets verify its high prediction accuracy. In future, we plan to investigate an ensemble model with various Swarm Intelligence-related LFA models, which can dynamically schedule these models considering various situations or datasets.